\documentclass{article}

\usepackage[preprint]{neurips_2026}

\usepackage[utf8]{inputenc}
\usepackage{amsmath}
\usepackage[T1]{fontenc}
\usepackage{hyperref}
\usepackage{url}
\usepackage{booktabs}
\usepackage{multirow}
\usepackage{makecell}
\usepackage{amsfonts}
\usepackage{amssymb}
\usepackage{nicefrac}
\usepackage{microtype}
\usepackage{xcolor}
\usepackage{algorithm}
\usepackage{algorithmic}

\usepackage{graphicx}

\title{Need We Teach Foundation Models What is a Generative Image? \\ \vspace{0.5em}
 \large Gradient-Free Generative Artifact Detection via Analytic Spectral Adaptation}

\author{
  Qiaoyu Chen \\
  Harbin University of Commerce \\
  \texttt{qiaoyu.chen.cs@gmail.com} \\
  \And
  Bing Zhang \\
  Harbin University of Commerce \\
  \texttt{zhangb@hrbcu.edu.cn} \\
}

\begin{document}

\maketitle

\begin{abstract}
Adapting foundation models to detect generative artifacts via gradient-based parameter updates inherently compromises their intrinsic representation capability.
During optimization, biased by a limited number of new samples, models tend to overfit to local domain shortcuts.
Specifically, fine-tuning massive pre-trained weights on limited specialized samples introduces erroneous inductive biases, inducing a measurable $\mathcal{L}_2$ norm perturbation in the high dimensional feature space---a phenomenon we formalize as \textit{anchor drift}.
Amplified subsequently via nonlinear activation functions, this drift severely impairs the model's zero-shot forgery detection across unseen domains.
To break this bottleneck, we propose a completely gradient free methodology that reframes detection from discriminative binary classification to an out-of-distribution (OOD) anomaly measurement problem.
By treating a frozen foundation model as a stable coordinate system, we establish an absolute natural anchor on the real visual manifold by analytically decoupling statistical and semantic deviations, derived respectively from attention-weighted spatial moments and perceptual inconsistency measurements of pristine images via orthogonal projection.
Evaluated within a unified, extreme zero-shot framework which restricts the source domain exclusively to classical face forgeries while testing on universal Text-to-Image (T2I) generations, our method outperforms gradient optimized paradigms by substantial margins, establishing unprecedented domain robustness.
Crucially, our reliance on backpropagation free forward pass features and linear solvers enables hardware agnostic, edge deployable calibration at significantly reduced latency.
Furthermore, the Sherman-Morrison formula unlocks instantaneous online learning to dynamically adapt against novel attacks, and inherently enables privacy preserving federated collaboration via covariance delta transmission.
Ultimately, we demonstrate that in rapidly evolving adversarial environments, \textit{agility is the ultimate robustness}.
\end{abstract}

\begin{figure}[t]
    \centering
    \includegraphics[width=\textwidth, trim={2cm 0 0 0}]{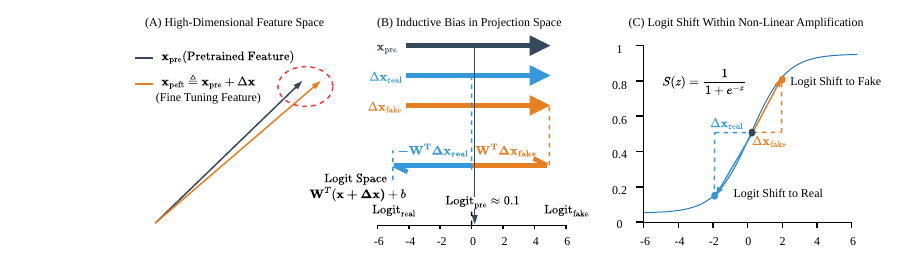}
    \caption{\textbf{Anchor drift phenomenon in parameter optimization.} Parameter optimization introduces an $\mathcal{L}_2$ norm discrepancy ($\|f_{\mathrm{fine-tuned}}(x)\|_2 \neq \|f_{\mathrm{pretrained}}(x)\|_2$) in the feature space (A). This inductive bias manifests in logit outputs (B) and is exponentially amplified in the probability space under nonlinear scaling, causing severe displacement of the decision boundary (C).}
    \label{fig:collapse}
\end{figure}
\vspace{-3mm}

\section{Introduction}
The rapid advancement of generative AI has led to the proliferation of high-fidelity synthetic content,
posing significant information security risks and challenging digital trust.
Recent foundational advancements in generative synthesis, ranging from latent diffusion models~\citep{rombach2022high}, Score-based SDEs~\citep{song2021score} and flow-matching frameworks~\citep{lipman2023flow} to classical autoregressive models and GANs~\citep{goodfellow2014generative, karras2019style, razavi2019generating, peebles2023scalable}, have vastly outpaced traditional detection mechanisms. While evaluating on unified testbeds like DeepfakeBench~\citep{yan2023deepfakebench}, robust detection mechanisms are imperative.

While the artifact detection paradigm has evolved from full fine-tuning (FFT) \citep{rossler2019faceforensics++,qian2020thinking, xu2021consistent,ni2022core,yan2023ucf,wang2020cnn} to parameter-efficient fine-tuning (PEFT) \citep{hu2022lora,yan2025orthogonal,cui2025forensics,yermakov2026deepfake} to preserve pre-trained knowledge, pushing this logic to its extreme exposes a paradox: how can a model robustly learn the evolving manifold of generative artifacts while relying on a limited, biased set of synthetic samples?

To break this paradox, we propose a fundamentally different paradigm. Rather than using pre-trained weights as a fragile starting point for optimization, we treat the frozen foundation model as a stable coordinate system that encapsulates the physics and semantics of the natural world. Within this space, we establish a natural anchor by extracting statistical moments and orthogonal bases exclusively from pristine images. Our method then analytically measures the discrepancy between any query image and this anchor---reframing generative artifact detection not as binary classification, but as a gradient-free anomaly measurement process.

To operationalize this theoretical insight, we introduce a novel, gradient-free generative artifact detection framework via Analytic Spectral Adaptation. Our main contributions are summarized as follows:
\begin{itemize}
\item \textbf{Anchor drift phenomenon in parameter optimization}: We formalize the phenomenon of \textit{anchor drift}, demonstrating that updating pre-trained weights introduces erroneous inductive biases and induces a measurable $\mathcal{L}_2$ norm perturbation in the high dimensional feature space. Amplified subsequently via nonlinear activations, this perturbation forces the decision boundary to systematically collapse toward the training manifold, resulting in a severe degradation of zero shot forgery detection across unseen domains.
\item \textbf{Attention-guided autofocus mechanism}: We propose an attention-guided mechanism leveraging the \texttt{[CLS]} token to automatically focus on salient objects within the image. This approach naturally bypasses the need for explicit spatial alignment or pose normalization, extracting purified feature maps with a high signal-to-noise ratio (SNR).
\item \textbf{Untrusted Layers Assumption}: We challenge the conventional paradigm that implicitly trusts all pre-trained representations. We posit that not all ViT blocks provide a positive SNR for forgery detection. Consequently, explicitly identifying and pruning these detrimental layers is essential to prevent noise accumulation and maximize discriminative capacity.
\end{itemize}

We name our framework \textbf{Lightning}---reflecting its lightweight, instantaneous calibration and the ability to strike directly at the essence of generative artifacts.

\section{Methodology}

\subsection{In-Method Regularization and In-Domain Overfitting}
While PEFT mitigates the severe in-method overfitting of FFT by restricting model capacity, it fundamentally remains a gradient-based optimization process. Forced to minimize the loss on a highly limited source distribution, the model inevitably resorts to shortcut learning \citep{geirhos2020shortcut}, memorizing domain-specific spurious correlations.

This shortcut learning introduces erroneous inductive biases, causing feature representations to deviate from the robust natural manifold---a phenomenon we term \textit{Anchor Drift} (evidenced by the growing $\mathcal{L}_2$ norm discrepancy in Figure~\ref{fig:collapse}). To formally understand this amplification, consider the pre-classification logit $z = w^\top f(x) = \|w\|_2 \|f(x)\|_2 \cos(\theta)$. While PEFT successfully aligns the feature direction $\theta$, the optimization distorts the feature magnitude $\|f(x)\|_2$ to fit source-domain shortcuts. During cross-domain inference, this magnitude distortion is exponentially amplified by Softmax in the probability space, displacing the decision boundary---a mechanism distinct from the linear probe analysis of \citet{kumar2022fine}.

\subsection{Butterfly Effect: The Cascading Effect of Gradient Noise}
\label{sec:butterfly}
Both LayerNorm-tuning and low-rank adapters share a common mechanism: modulating the contribution of each pre-trained block to the main residual stream. We pose a pivotal issue: \textit{does every pre-trained ViT block truly provide positive gain for forgery detection?} Given the vast domain gap between natural image pre-training and generative artifact detection, the necessity of every layer remains an open question. We argue that because gradient-based scaling parameters can never rigorously converge to absolute zero, both methods fail to explicitly block forward signal propagation from detrimental layers. This triggers a butterfly effect: untreated noise inevitably leaks into the residual stream, cascading and amplifying across the network depth.

\textbf{Remark 1 (The Limits of Soft Suppression).} We posit that soft suppression via gradient updates fails to disable detrimental layers. Specifically, the violent scaling in LayerNorm operations forces low-variance noise to compete with discriminative signals, while the non-convex nature of deep learning optimization prevents scaling parameters ($\gamma_l$) from converging to absolute zero. Consequently, untreated noise inevitably leaks and accumulates across the network depth, fundamentally compromising the discriminative boundary for generative artifacts. (Detailed mathematical proof is deferred to Appendix~\ref{app:soft_suppression}).

\begin{figure}[t]
    \centering
    \includegraphics[width=\textwidth]{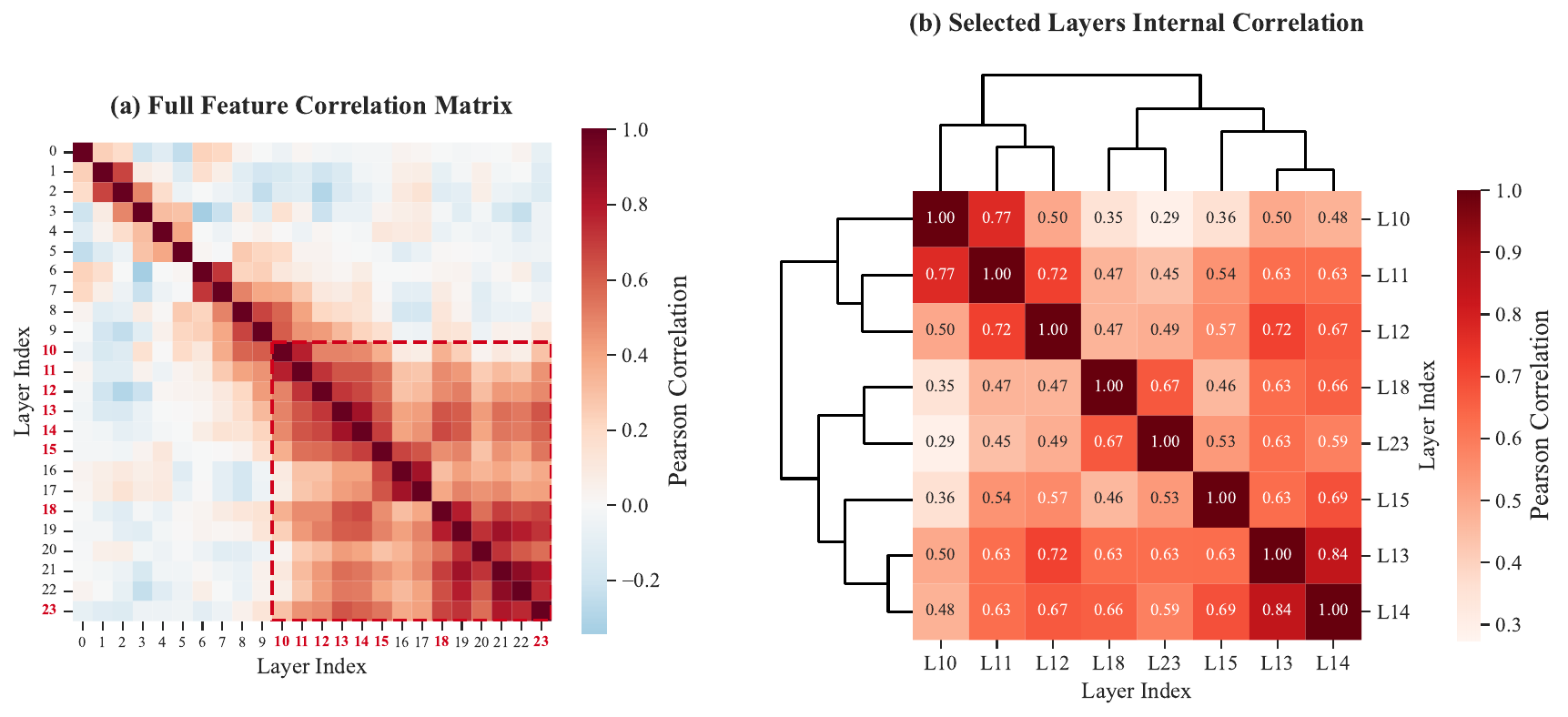}
    \caption{\textbf{Correlation map of layer-wise discriminative signals in CLIP (ViT-L/24).} Evaluated via per-layer Ridge Regression, predictive signals across different layers exhibit severe inconsistency. This heterogeneity confirms that the unpruned space embeds conflicting representations, justifying our targeted layer selection. (Note: While this visualization isolates per-layer probes to expose raw inconsistencies, our selection algorithm subsequently employs a synergistic global probe).}
    \label{fig:correlation}
\end{figure}

\subsection{Occam's Razor: Not Just Layers, But Dimensions}
To actualize the explicit, hard layer pruning necessitated in Section~\ref{sec:butterfly} while averting the curse of dimensionality during layer evaluation, our dual-pruning strategy mandates that we first drastically compress task-relevant representations within each layer (Dimensions) prior to conducting a global layer search (Layers).

\textbf{Orthogonal Decoupling via Attention-Guided Probing.} Rather than relying on monolithic representations, we analytically decouple the pre-trained features into a statistical manifold $x_{\mathrm{stat}} = [ \mu \parallel \sigma ] \in \mathbb{R}^{2048}$ capturing natural fidelities, and a semantic manifold $x_{\mathrm{sem}} = [ \mathrm{consist} \parallel \mathrm{gap} ] \in \mathbb{R}^{2048}$ capturing perceptual consistency, where $\mathrm{consist}$ isolates high-frequency texture anomalies without requiring heuristic pose alignments. To bypass explicit spatial alignment, we utilize the normalized \texttt{[CLS]} token as a semantic query to compute an intrinsic attention map over spatial tokens. This attention map strictly guides the extraction of moments ($\mu, \sigma$), the semantic residual ($\mathrm{gap}$), and high-frequency texture anomalies ($\mathrm{consist}$). Detailed mathematical formulations demonstrating how texture filtering and attention routing bypass explicit alignment are deferred to Appendix~\ref{app:texture_filtering}.

Subsequently, we equilibrate individual dimensions via Z-score standardization ($\tilde{x}_{s} = (x_{s} - \mu_{\mathrm{global},s}) / \sqrt{\sigma_{\mathrm{global},s}^2 + \epsilon}$) where $\sigma_{\mathrm{global},s}^2$ denotes variance, and construct the empirical covariance $\Sigma_s = \tilde{X}_s^T \tilde{X}_s$. Crucially, deriving the natural manifold poses a profound robustness challenge against rank-deficiency. In extreme few-shot scenarios ($N \ll D$), $\Sigma_s$ becomes severely ill-conditioned, collapsing standard eigendecomposition. Leveraging the mathematical equivalence between eigendecomposition and SVD \citep{halko2011finding}, we directly apply SVD to $\tilde{X}_s$ to robustly extract the top-$K$ right singular vectors $V_{K,s}$. This circumvents direct covariance inversion and is numerically equivalent to PCA via the pseudoinverse of $\Sigma_s$. Projecting features onto these bases aggressively compresses each decoupled space from 2048D to $K=64$, isolating highly discriminative subspaces while fundamentally immunizing against ill-posed matrix collapse.

\textbf{Robust Layer Search via Decoupled Global Ridge Probing.} As motivated by the severe inconsistency observed in layer-wise signals (Figure~\ref{fig:correlation}), selecting the optimal layer subset $\mathcal{S}$ is essential. To isolate this subset without suffering severe collinearity from naive concatenation, we employ a global Ridge Regression \citep{hoerl1970ridge} probe. Unlike per-layer probing, this naturally captures inter-layer synergies. We quantify each layer's discriminative contribution by computing the $L_2$ norm of its corresponding block-wise regression weights (defining its spectral energy). By aggregating these statistical and semantic energies via bagging and weighting coefficients ($\alpha, \beta$), we robustly select the top-$K$ synergistic layers (formalized in Algorithm~\ref{alg:rlp_sea}).
\begin{algorithm}[h]
\caption{Robust Layer Pruning via Spectral Energy Analysis (RLP-SEA)}
\label{alg:rlp_sea}
\begin{algorithmic}[1]
\REQUIRE Frozen ViT with $L$ layers, Calibration data $\mathcal{D} = \{(X_i, y_i)\}_{i=1}^N$, Target layers $K$, Bagging rounds $R$, Weighting coefficients $\alpha, \beta$, Ridge penalty $\lambda$.
\ENSURE Optimal layer subset $\mathcal{S}$ of size $K$.
\STATE \textbf{Extract \& Compress:} For all $l \in \{1 \dots L\}$, extract and orthogonally project features (via SVD to dimension $K$) to obtain decoupled representations $X_{\mathrm{stat}}^{(l)}, X_{\mathrm{sem}}^{(l)}$. Concatenate to form $X_{\mathrm{stat}}, X_{\mathrm{sem}} \in \mathbb{R}^{N \times (L \times K)}$.
\STATE Initialize accumulated layer energy $E = \mathbf{0} \in \mathbb{R}^L$.
\FOR{$r = 1$ \TO $R$}
    \STATE Sample bagged subset $(\tilde{X}, \tilde{y}) \sim \mathcal{D}$ via bootstrap sampling.
    \STATE $W_{\mathrm{stat}} \leftarrow (\tilde{X}_{\mathrm{stat}}^T \tilde{X}_{\mathrm{stat}} + \lambda I)^{-1} \tilde{X}_{\mathrm{stat}}^T \tilde{y}$ \hfill \textit{// Global Ridge Probing}
    \STATE $W_{\mathrm{sem}} \leftarrow (\tilde{X}_{\mathrm{sem}}^T \tilde{X}_{\mathrm{sem}} + \lambda I)^{-1} \tilde{X}_{\mathrm{sem}}^T \tilde{y}$
    \FOR{$l = 1$ \TO $L$}
        \STATE Extract block-wise weights: $W_{\mathrm{stat}}^{(l)} \leftarrow W_{\mathrm{stat}}[(l-1)K : lK]$
        \STATE Extract block-wise weights: $W_{\mathrm{sem}}^{(l)} \leftarrow W_{\mathrm{sem}}[(l-1)K : lK]$
        \STATE $e_{\mathrm{stat}}^{(l)} = \|W_{\mathrm{stat}}^{(l)}\|_2, \quad e_{\mathrm{sem}}^{(l)} = \|W_{\mathrm{sem}}^{(l)}\|_2$ \hfill \textit{// Layer-wise spectral energy}
    \ENDFOR
    \STATE Normalize $e_{\mathrm{stat}}, e_{\mathrm{sem}}$ independently; \quad $E \leftarrow E + (\alpha \cdot e_{\mathrm{stat}} + \beta \cdot e_{\mathrm{sem}})$
\ENDFOR
\STATE Average energy: $E \leftarrow E / R$.
\RETURN $\mathcal{S} \leftarrow \mathrm{Indices\ of\ Top-}K \mathrm{\ values\ in\ } E$.
\end{algorithmic}
\end{algorithm}

\subsection{Semantic Soft-Routing for Unpaired Alignment}
Inspired by activation steering in NLP \citep{turner2024activation}, where manipulating latent directional signals controls model behavior without weight updates, we recognize that generative artifact detection is fundamentally a signal-to-noise ratio (SNR) problem: as long as the steering direction correctly isolates the forgery manifold, weight projection suffices---rendering gradient-based optimization unnecessary. Formally, discovering generative artifacts is akin to identifying an Activation Steering Vector in the latent space---a discriminative direction pointing from the pristine manifold to the manipulated distribution. Traditionally, establishing such a directional vector requires strict $(x_{\mathrm{real}}, x_{\mathrm{fake}})$ image pairs, a luxury exclusively available in localized manipulation datasets (e.g., Face Swapping) where the original source image is known. However, modern advanced generators (e.g., Diffusion and Flow-matching models) synthesize images entirely from noise, rendering them inherently unpaired and severely complicating the extraction of purely artifact-driven steering directions.

To overcome this, we propose a Semantic Soft-Routing mechanism to construct robust \textit{virtual} pairs during the calibration phase. For any isolated fake sample $x_f$ in the calibration set, we perform K-nearest neighbor routing within a diverse pristine image pool in the derived semantic and statistical subspaces using cosine similarity to identify its closest stylistic and semantic matches. By applying a temperature-scaled softmax over their similarity scores, we blend these neighbors to synthesize a ``soft real'' anchor $x_r^*$. This virtual pairing forcibly aligns the content and style distributions between the true natural manifold and the synthesized artifact space. Consequently, during calibration, the solver is compelled to ignore trivial intra-class variations (e.g., lighting, poses) and strictly focus on the residual difference $(x_f - x_r^*)$, establishing a universally robust steering boundary even against entirely unpaired full-image generations. Crucially, this routing is executed exclusively during calibration to derive the optimal steering weights $w_s$; during inference, the model achieves near-zero latency by simply performing linear dot products via the pre-computed weights without any KNN retrieval.

\subsection{Analytic Calibration and Online Adaptation}
\label{sec:analytic_calibration}
To construct the final calibrator, we concatenate the compressed features across the optimal layer subset $\mathcal{S}$:
\begin{equation}
    X_{\mathrm{global},s} = \left[ (\tilde{X}_{1,s} V_{K,1,s}) \parallel \dots \parallel (\tilde{X}_{|\mathcal{S}|,s} V_{K,|\mathcal{S}|,s}) \right] \in \mathbb{R}^{N \times (|\mathcal{S}| \times K)}
\end{equation}
This concatenation functions as a layer-wise ensemble voting mechanism, ensuring anomaly detection relies on robust cross-layer consensus. We then solve a Ridge Regression classifier:
\begin{equation}
    w_s = (X_{\mathrm{global},s}^T X_{\mathrm{global},s} + \lambda I)^{-1} X_{\mathrm{global},s}^T y
\end{equation}
which unifies implicit Mahalanobis whitening via the empirical covariance precision matrix and extracts a precision-scaled steering vector $w_s \propto \Sigma_s^{-1}(\mu_1 - \mu_0)$. During inference, predictions from orthogonal manifolds are late-fused via element-wise maximum selection. Under extreme rank deficiency, we fall back to the SVD pseudoinverse. Detailed equivalence proofs are in Appendix~\ref{app:ridge_equivalence}.

\textbf{Recursive Least Squares (RLS) for Anti-Forgetting Online Learning.}
A critical advantage of our analytic formulation is that it naturally instantiates a Recursive Least Squares (RLS) estimator~\citep{plackett1950some, haykin2002adaptive}, which provides \textit{mathematically exact} continual learning without catastrophic forgetting. Upon encountering a new sample $(x_{t+1}, y_{t+1})$, the precision matrix $P_t = (X_t^T X_t + \lambda I)^{-1}$ and weights $w_t$ are updated via rank-one Sherman-Morrison modifications:
\begin{align}
    P_{t+1} &= P_t - \frac{P_t x_{t+1} x_{t+1}^T P_t}{1 + x_{t+1}^T P_t x_{t+1}} \\
    w_{t+1} &= w_t + P_{t+1} x_{t+1} (y_{t+1} - x_{t+1}^T w_t)
\end{align}
This yields a solution \textit{identical} to batch retraining on the full history $\{X_{1:t+1}, y_{1:t+1}\}$, guaranteeing zero forgetting by construction---no replay buffers, no regularization penalties, and no approximation errors. The RLS framework is the optimal linear estimator in the sequential setting, achieving the exact least-squares solution at each timestep with $\mathcal{O}(d^2)$ per-update cost.

\textbf{Covariance for Offline Batch Learning.}
Conversely, when the full calibration set is available offline, the batch Ridge solution operates directly on the empirical covariance $\Sigma = X^T X$. This covariance matrix provides a complete second-order characterization of the feature distribution, enabling principled uncertainty quantification and regularization via the spectral decomposition $\Sigma = V \Lambda V^T$. The resulting estimator $w = (X^T X + \lambda I)^{-1} X^T y$ is the minimum-variance unbiased linear estimator under Gaussian noise (Gauss-Markov theorem \citep{plackett1950some}), providing rigorous statistical guarantees. Together, the RLS online mode and covariance-based batch mode form a unified analytic ecosystem with provable optimality, exact knowledge preservation, and zero gradient-induced anchor drift.

\textbf{Federated Covariance Aggregation.}
In decentralized settings, the global covariance decomposes additively: $(X^T X)_{\mathrm{global}} = \sum_{i=1}^M X_i^T X_i$. Clients transmit only $K \times K$ covariance matrices ($\sim$8 KB in FP16 for $K=64$), enabling privacy-preserving collaboration without raw data sharing. Detailed discussion is deferred to Appendix~\ref{app:federated}.

\subsection{Decoupled Augmentation for Anomaly Manifold Calibration}
Unlike deep neural networks, rigid statistical solvers are susceptible to covariance pollution from augmented samples. To circumvent this, we isolate augmented samples to construct an independent ``Augmented Anchor'', measuring the Mahalanobis distance $d$ of any query to the natural manifold and applying an exponential decay penalty $\sigma = 1 + \alpha \exp(-\gamma \cdot d^2)$ during inference. This transforms augmentation into a safe anomaly penalty. Full details are in Appendix~\ref{app:decoupled_aug}.

\section{Experiments}

\subsection{Implementation Details}
\label{sec:impl_details}
We employ the pre-trained CLIP ViT-L/14 as the frozen backbone. The extracted statistical and semantic features are independently reduced to $K=64$ dimensions via randomized SVD. RLP-SEA identifies an optimal subset of 8 layers out of 24. Hyperparameters are conservatively set: Ridge regularization $\lambda=0.15$, temperature $\tau=40.0$ for Semantic Soft-Routing, and $30\%$ masking with Gaussian noise ($\sigma=0.5$) for Decoupled Augmentation. These hyperparameters are fixed across all experiments unless otherwise specified. Crucially, to ensure a strict zero-shot evaluation protocol, all hyperparameters such as the intrinsic dimensionality $K=64$ and the selected layer subset $\mathcal{S}$ are determined exclusively on the source domain validation set. This rigorous setup effectively prevents potential target-domain data leakage.

\subsection{Zero-Shot Cross-Domain Evaluation on OpenFake}

\label{sec:openfake}

\paragraph{Evaluation Benchmarking.} To provide a comprehensive benchmark for comparison, we introduce 8 competitive detectors, including several classical detection methods such as Xception \cite{rossler2019faceforensics++} (ICCV'19), F3Net \cite{qian2020thinking} (ECCV'20), and CORE \cite{ni2022core} (CVPRW'22), and also several latest SOTA methods (after 2023), such as UCF \cite{yan2023ucf} (ICCV'23), ProDet \cite{cheng2024can} (NeurIPS'24), Effort \cite{yan2025orthogonal} (ICML'25), ForAda \cite{cui2025forensics} (CVPR'25), and GenD \cite{yermakov2026deepfake} (WACV'26). All detectors are trained or calibrated on FF++ at C23 quality.

\paragraph{Evaluation Protocol.} We adopt a strict zero-shot protocol where models are calibrated (Lightning) or trained (gradient-based baselines) on FaceForensics++ (FF++)~\citep{rossler2019faceforensics++} at C23 quality and tested on OpenFake~\citep{livernoche2025openfake}, a large-scale benchmark comprising diverse Text-to-Image (T2I) architectures (e.g., Diffusion, Flow-matching, and Proprietary engines), without any adaptation. Following \citet{shiohara2022detecting}, we sample 8 frames per video for calibration and 32 for inference; for the single-image OpenFake benchmark, we report per-frame AUC and AP.

Table~\ref{tab:openfake_per_gen} reports the per-generator AUC comparison on OpenFake for the 12 representative generators. Due to space constraints, the corresponding comprehensive Balanced AP breakdown is deferred to Appendix~\ref{app:ap_summary}, which exhibits identical superiority trends. The complete per-generator results across all 34 generators are provided in Table~\ref{tab:app_openfake_auc}, Table~\ref{tab:app_openfake_auc_2}, Table~\ref{tab:app_openfake_ap}, and Table~\ref{tab:app_openfake_ap_2} of Appendix~\ref{app:full_results}. Lightning achieves the highest average AUC (0.820) and Bal\_AP (0.799), surpassing the best gradient-based baseline (GenD, 0.618 AUC and 0.590 Bal\_AP) by significant margins. This gap empirically validates our core theoretical claim: parameter optimization on limited source-domain samples induces anchor drift, causing the decision boundary to collapse toward the FF++ manifold. In contrast, Lightning's frozen backbone preserves the CLIP representation space intact, enabling robust zero-shot generalization to unseen generative architectures. Notably, while previous gradient-based methods experience performance variations across unseen generative architectures, Lightning maintains consistent generalization capabilities. It performs well not only on classical latent-diffusion variants (e.g., SD-1.5 to SD-XL), but also across emerging flow-matching models (e.g., Flux.1) and proprietary engines (e.g., Midjourney v7). This architectural robustness suggests that the analytically decoupled manifold effectively captures intrinsic structural inconsistencies shared among diverse generative artifacts.

\begin{table}[h]
\centering
\caption{\textbf{Per-generator AUC (\%) comparison on OpenFake.} All methods are trained or calibrated on FF++ (C23) and tested zero-shot on OpenFake without any domain adaptation. Lightning reports the best calibration run. Best results in \textbf{bold}.}
\label{tab:openfake_per_gen}
\resizebox{\textwidth}{!}{
\begin{tabular}{l|c|cccc|ccc|ccccc|c}
\toprule
\multicolumn{1}{c|}{\multirow{2}{*}{\textbf{Method}}} & \multirow{2}{*}{\makecell{\textbf{Trainable}\\ \textbf{Params}}} & \multicolumn{4}{c|}{\textbf{Stable Diffusion}} & \multicolumn{3}{c|}{\textbf{Proprietary}} & \multicolumn{5}{c|}{\textbf{Flow \& Emerging}} & \multirow{2}{*}{\textbf{Avg}} \\
\cmidrule(lr){3-6} \cmidrule(lr){7-9} \cmidrule(lr){10-14}
 & & \textbf{Sd-1.5} & \textbf{Sd-2.1} & \textbf{Sd-3.5} & \textbf{Sdxl} & \textbf{Mj-v6} & \textbf{Mj-v7} & \textbf{Dall-e 3} & \textbf{Flux.1} & \textbf{Flux-real} & \textbf{Ideogram} & \textbf{Recraft} & \textbf{Chroma} & \\
\midrule
Xception \cite{rossler2019faceforensics++} & 83M & 0.561 & 0.590 & 0.634 & 0.514 & 0.536 & 0.599 & 0.565 & 0.563 & 0.529 & 0.535 & 0.573 & 0.638 & 0.570 \\
F3Net \cite{qian2020thinking} & 22M & 0.348 & 0.492 & 0.589 & 0.473 & 0.446 & 0.555 & 0.463 & 0.585 & 0.567 & 0.474 & 0.575 & 0.585 & 0.513 \\
CORE \cite{ni2022core} & 22M & 0.312 & 0.450 & 0.523 & 0.514 & 0.405 & 0.578 & 0.455 & 0.621 & 0.615 & 0.527 & 0.517 & 0.550 & 0.506 \\
UCF \cite{yan2023ucf} & 47M & 0.421 & 0.498 & 0.595 & 0.523 & 0.454 & 0.475 & 0.418 & 0.609 & 0.580 & 0.474 & 0.523 & 0.596 & 0.514 \\
ProDet$^\dagger$ \cite{cheng2024can} & 96M & 0.542 & 0.534 & 0.538 & 0.566 & 0.540 & 0.660 & 0.462 & 0.609 & 0.584 & 0.521 & 0.625 & 0.585 & 0.564 \\
Effort$^\dagger$ \cite{yan2025orthogonal} & 0.19M & 0.718 & 0.739 & 0.521 & 0.635 & 0.518 & 0.587 & 0.431 & 0.440 & 0.468 & 0.510 & 0.580 & 0.534 & 0.557 \\
ForAda$^\dagger$ \cite{cui2025forensics} & 5.7M & 0.717 & 0.774 & 0.518 & 0.631 & 0.424 & 0.505 & 0.393 & 0.448 & 0.468 & 0.495 & 0.461 & 0.584 & 0.535 \\
GenD$^\dagger$ \cite{yermakov2026deepfake} & 0.03M & 0.799 & 0.835 & 0.592 & 0.711 & 0.532 & 0.629 & 0.665 & 0.475 & 0.472 & 0.567 & 0.543 & 0.599 & 0.618 \\
\midrule
\textbf{Lightning (Ours)} & \textbf{0M} & \textbf{0.800} & \textbf{0.905} & \textbf{0.844} & \textbf{0.886} & \textbf{0.693} & \textbf{0.864} & \textbf{0.706} & \textbf{0.855} & \textbf{0.830} & \textbf{0.816} & \textbf{0.748} & \textbf{0.887} & \textbf{0.820} \\
\bottomrule
\end{tabular}
}
\end{table}
\vspace{-3mm}

\subsection{Cross-Benchmark Transfer Cost Analysis}

\label{sec:ufd_transfer}

To quantify the \textit{transfer cost}---the performance degradation incurred when a model calibrated on one domain is deployed on another---we conduct a bidirectional evaluation between FF++ and UniversalFakeDetect (UFD)~\citep{ojha2023towards}. Unlike the per-generator breakdown in Section~\ref{sec:openfake}, we report global AUC and AP aggregated across all generators, directly reflecting the practical deployment scenario where the generator identity is unknown.

We define the \textit{Robust Transfer Index} (RTI) to holistically quantify transfer robustness---penalizing both low absolute performance and severe cross-domain degradation. Unlike a naive stability ratio which spuriously rewards models lacking capacity, RTI explicitly scales the absolute zero-shot capability by its preservation ratio:
\begin{equation}
    \text{RTI} = \frac{(\text{Metric}_{\text{zero-shot}})^2}{\text{Metric}_{\text{fine-tuned}}}
\end{equation}
A higher RTI indicates that the model not only achieves an exceptionally high performance ceiling but also securely preserves it without requiring target-domain adaptation. Table~\ref{tab:transfer_cost} reports the zero-shot and fine-tuned results on UFD, along with the computed RTI. Lightning achieves an RTI of 0.854 in AUC and 0.872 in AP, demonstrating that its frozen backbone preserves the CLIP representation space so effectively that fine-tuning on the target domain yields marginal improvement. For instance, although parameter-efficient baselines (e.g., GenD) achieve competitive fine-tuned metrics by updating merely a fraction of weights, their RTI remains constrained by limited zero-shot retention. This aligns with our core methodology: forgery detection is fundamentally a Signal-to-Noise Ratio (SNR) problem. By treating the powerful foundation model as a strictly frozen, stable coordinate system and entirely abandoning gradient updates, Lightning explicitly prevents the noise accumulation inherent in parameter tuning, thereby preserving pristine zero-shot capabilities.

\begin{table}[htbp]
\centering
\caption{\textbf{Cross-benchmark transfer robustness on UniversalFakeDetect.} The zero-shot evaluation applies models utilizing their official FF++ pre-trained weights directly to UFD test samples. The fine-tuned evaluation reports the performance upper bound achieved by re-optimizing the models locally on the UFD training split.}
\label{tab:transfer_cost}
\resizebox{\textwidth}{!}{
\begin{tabular}{l|c|c|c|c|c|c}
\toprule
\multirow{2}{*}{\textbf{Method}} & \multicolumn{2}{c|}{\textbf{Zero-shot}} & \multicolumn{2}{c|}{\textbf{Fine-tuned}} & \multicolumn{2}{c}{\textbf{RTI}} \\
\cmidrule(lr){2-3} \cmidrule(lr){4-5} \cmidrule(lr){6-7}
 & \textbf{AUC} & \textbf{AP} & \textbf{AUC} & \textbf{AP} & \textbf{AUC} & \textbf{AP} \\
\midrule
Xception \cite{rossler2019faceforensics++} & 0.562 & 0.630 & 0.881 & 0.896 & 0.359 & 0.442 \\
F3Net \cite{qian2020thinking} & 0.619 & 0.667 & 0.824 & 0.850 & 0.466 & 0.524 \\
CORE \cite{ni2022core} & 0.551 & 0.601 & 0.857 & 0.891 & 0.354 & 0.405 \\
UCF \cite{yan2023ucf} & 0.595 & 0.630 & 0.837 & 0.860 & 0.423 & 0.461 \\
ProDet \cite{cheng2024can} & 0.588 & 0.615 & 0.819 & 0.848 & 0.422 & 0.446 \\
Effort \cite{yan2025orthogonal} & 0.833 & 0.869 & 0.966 & \textbf{0.978} & 0.718 & 0.772 \\
ForAda \cite{cui2025forensics} & 0.774 & 0.817 & \textbf{0.968} & 0.977 & 0.619 & 0.683 \\
GenD \cite{yermakov2026deepfake} & 0.803 & 0.828 & 0.950 & 0.966 & 0.678 & 0.710 \\
\midrule
\textbf{Lightning (Ours)} & \textbf{0.903} & \textbf{0.918} & 0.955 & 0.966 & \textbf{0.854} & \textbf{0.872} \\
\bottomrule
\end{tabular}
}
\end{table}
\vspace{-3mm}

\subsection{Analytical Agility and Pareto Optimality}
\label{sec:agility_pareto}

To establish a hardware-agnostic measure of training efficiency, we quantify the total computational cost via \emph{Total Training FLOPs} (floating-point operations). For gradient-based methods, this accounts for both forward and backward propagation: $\text{Total} = F_{\text{fwd}} \times N \times (1 + \alpha)$, where $\alpha \approx 2$ reflects the typical cost of automatic differentiation. For gradient-free methods, no backward cost is incurred.

\begin{figure}[t]
\centering
\includegraphics[width=\linewidth]{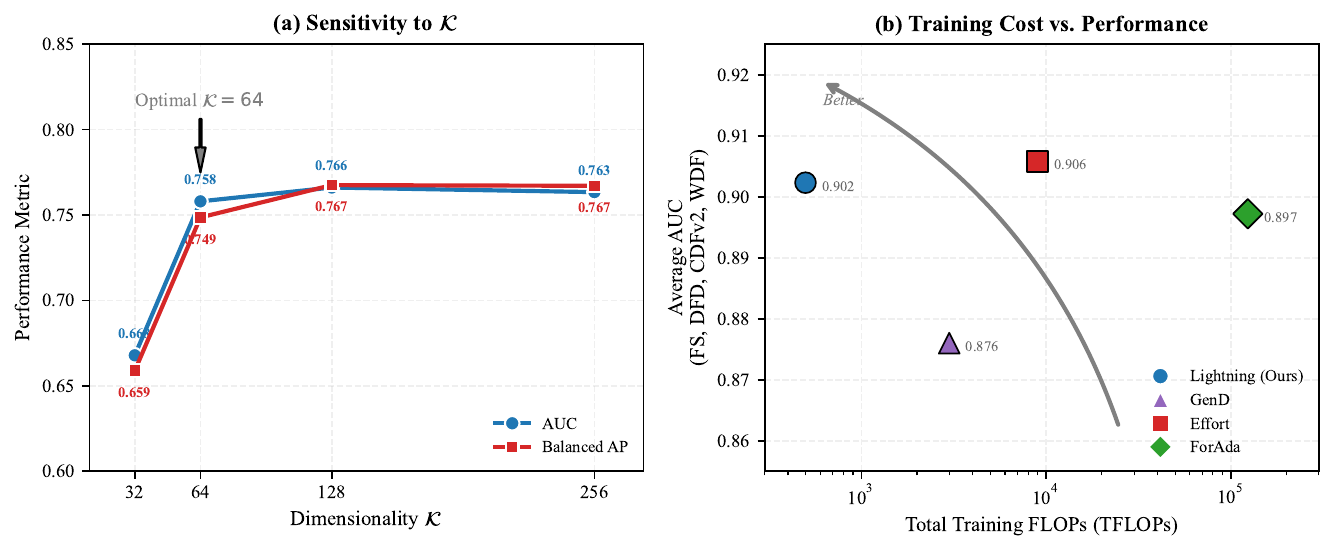}
\caption{\textbf{Efficiency and feature-space analysis.}
\textbf{(a)} Sensitivity to anchor dimensionality $\mathcal{K}$.
AUC and Balanced AP saturate at $\mathcal{K}=64$, confirming that a compact feature space ($64$ dimensions) captures nearly all discriminative signal with minimal computational overhead.
\textbf{(b)} Pareto analysis of training cost vs.\ detection performance.
Lightning (ours, blue circle, $498$\,TFLOPs, avg.\ AUC $0.902$),
GenD~\citep{yermakov2026deepfake} (purple triangle, $2\,989$\,TFLOPs, $0.876$),
Effort~\citep{yan2025orthogonal} (red square, $8\,978$\,TFLOPs, $0.906$),
and ForAda~\citep{cui2025forensics} (green diamond, $123\,780$\,TFLOPs, $0.897$).
The arrow indicates the desired direction (lower cost, higher AUC).
All AUCs are video-level averages over FaceShifter~\citep{li2019faceshifter}, DFD~\citep{google2019deepfakedetection}, CelebDFv2~\citep{li2020celeb}, and WildDeepFake~\citep{zi2020wilddeepfake}.
}
\label{fig:combined}
\end{figure}

Figure~\ref{fig:combined} presents two complementary analyses.
First, the sensitivity to the anchor dimensionality $\mathcal{K}$ (Fig.~\ref{fig:combined}a) reveals a clear saturation point: AUC rises sharply from $0.668$ at $\mathcal{K}=32$ to $0.758$ at $\mathcal{K}=64$, then plateaus near $0.765$ for $\mathcal{K}\ge128$.
This demonstrates that a remarkably compact $64$-dimensional subspace suffices to capture the discriminative anchor-drift signal, confirming that our method operates in a~\emph{computationally efficient and statistically robust} feature regime where performance converges at minimal representational cost.

Second, the Pareto landscape (Fig.~\ref{fig:combined}b) shows that Lightning operates within $498$ TFLOPs---merely $5.5\%$ of Effort's training budget---yet attains a competitive average AUC of $0.902$ versus $0.906$. More importantly, Lightning introduces \emph{zero trainable parameters} throughout the entire procedure; the backbone remains frozen, and the only operation is an analytical SVD-based anchor extraction followed by a closed-form ridge regression. In sharp contrast, Effort optimizes $7.5$M parameters via full forward-backward propagation, giving it substantial capacity to overfit to the training-domain distribution.

This distinction is critical: any method with learnable parameters can, in principle, inflate its in-domain performance by memorizing dataset-specific biases. Lightning, being entirely parameter-free, cannot do so. The fact that it still matches gradient-based counterparts therefore provides strong evidence that the anchor-drift signals we capture are \emph{intrinsically universal}---they reflect structural artifacts of generative models rather than domain-specific idiosyncrasies. ForAda~\citep{cui2025forensics}, despite employing a stronger CLIP ViT-L/14 backbone and $60$ training epochs, incurs over two orders of magnitude more FLOPs ($123\,780$ TFLOPs) without surpassing Lightning's AUC, further underscoring the structural efficiency advantage of gradient-free analytic alignment.

\section{Related Work}
Generative artifact detection has evolved from full fine-tuning (FFT) of CNNs~\citep{rossler2019faceforensics++, qian2020thinking, ni2022core} to parameter-efficient fine-tuning (PEFT)~\citep{hu2022lora, yan2025orthogonal, cui2025forensics, yermakov2026deepfake} to mitigate in-domain overfitting. However, both paradigms fundamentally rely on gradient-based optimization, which introduces anchor drift and catastrophic forgetting under domain shift~\citep{li2023continual}. Continual learning methods (EWC~\citep{kirkpatrick2017overcoming}, iCaRL~\citep{rebuffi2017icarl}) and federated approaches (FedAvg) further compound this with computational overhead and privacy concerns. Lightning breaks from all these paradigms by reframing detection as a gradient-free anomaly measurement problem, where the frozen backbone serves as an immutable coordinate system and the linear solver provides mathematically exact online/offline learning with zero forgetting.

\section{Conclusion}

In this work, we challenge the pervasive paradigm of gradient-based feature alignment in generative artifact detection. We demonstrate that parameter optimization inherently compromises the pristine representation space of foundation models, leading to anchor drift and catastrophic cross-domain generalization failures. To resolve this, we introduce Lightning, a fully gradient-free Analytic Spectral Adaptation framework. By establishing an absolute natural anchor via explicit semantic-statistical decoupling and robust orthogonal projection, Lightning reframes detection as an out-of-distribution measurement process. Evaluated on an extreme zero-shot protocol encompassing 34 diverse generative architectures, our method sets a new state-of-the-art in cross-domain robustness. Crucially, its reliance on linear solvers enables zero-latency calibration and instantaneous online adaptation without catastrophic forgetting. Ultimately, in rapidly evolving adversarial environments, analytic agility is the ultimate robustness.

\begin{ack}
This work was supported by the National Innovation Training Program for College Students (Grant No. 202410240135).
\end{ack}

\newpage
\bibliographystyle{abbrvnat}
\bibliography{refs}

\clearpage
\appendix

\section{Detailed Mathematical Formulations and Algorithms}
\label{app:detailed_methods}

\subsection{Texture Filtering and Semantic Gap Formulation}
\label{app:texture_filtering}
To construct the 2048D semantic representation $x_{\mathrm{sem}}$, we rigorously quantify the spatial artifact inconsistency. Let $f_{\mathrm{spatial}} \in \mathbb{R}^{B \times C \times H \times W}$ denote the spatial tokens extracted from the ViT backbone. We apply orthogonal Sobel-like convolution filters $\mathcal{K}_h$ and $\mathcal{K}_v$ to compute the horizontal and vertical gradients. The global texture baseline $T_{\mathrm{global}}$ is formulated as the spatial root-mean-square of these gradients across the entire image:
\begin{equation}
    T_{\mathrm{global}} = \sqrt{\frac{1}{H \cdot W} \sum_{i,j} \left( (f_{\mathrm{spatial}} * \mathcal{K}_h)_{i,j}^2 + (f_{\mathrm{spatial}} * \mathcal{K}_v)_{i,j}^2 \right)}
\end{equation}
Concurrently, the local texture anomaly $T_{\mathrm{local}}$ is extracted by weighting the squared gradients strictly with the intrinsic attention map $A$ derived from the normalized \texttt{[CLS]} token:
\begin{equation}
    T_{\mathrm{local}} = \sqrt{\frac{\sum_{i,j} A_{i,j} \cdot \left( (f_{\mathrm{spatial}} * \mathcal{K}_h)_{i,j}^2 + (f_{\mathrm{spatial}} * \mathcal{K}_v)_{i,j}^2 \right)}{\sum_{i,j} A_{i,j} + \epsilon}}
\end{equation}
The final spatial artifact inconsistency ($\mathrm{consist}$) is measured as the log-ratio: $\log(T_{\mathrm{local}} + \epsilon) - \log(T_{\mathrm{global}} + \epsilon)$. This explicitly isolates high-frequency structural artifacts within salient regions, bypassing the need for heuristic pose alignments.

\subsection{Mathematical Proof: The Limits of Soft Suppression}
\label{app:soft_suppression}
Let $x_l$ denote the input feature to the $l$-th transformer block, and $f_l$ denote the noisy sub-layer (e.g., capturing irrelevant background textures). Modern ViTs \citep[e.g., CLIP,][]{radford2021learning} employ a Pre-LayerNorm architecture, where the output is formulated as:
\begin{equation}
x_{l+1} = x_l + f_l\left( \gamma_l \cdot \frac{x_l - \mu_l}{\sqrt{\sigma_l^2 + \epsilon}} + \beta_l \right)
\end{equation}
where $\mu_l$ and $\sigma_l^2$ denote the mean and variance of the input $x_l$, and $\gamma_l, \beta_l$ are the learnable affine parameters.
In an end-to-end gradient-based fine-tuning paradigm (e.g., PEFT tuning LN params), the optimizer attempts to suppress this detrimental layer by driving the scaling factor $\gamma_l \to 0$. However, two fundamental constraints prevent true layer disabling:

\begin{enumerate}
    \item \textbf{Forced Noise Amplification}: As evidenced by the division by $\sqrt{\sigma_l^2 + \epsilon}$, the normalization operation equalizes the scale of all input dimensions. This effectively removes the natural suppression of low-variance background noise, forcing it to compete equally with discriminative signals. To counteract this violent scaling, the optimizer must aggressively push $\gamma_l$ toward zero.
    \item \textbf{Optimization Lower Bounds}: Due to the highly non-convex nature of deep learning optimization and conflicting gradient signals, $\gamma_l$ practically never converges to absolute zero.
\end{enumerate}

Consequently, the violently amplified noise multiplied by a stubbornly non-zero $\gamma_l$ (as gradient descent on non-convex objectives guarantees convergence only to a stationary point, not the global minimum of $\|\gamma_l\|$) ensures that the term $f_l(\gamma_l \cdot \mathrm{Norm}(x_l) + \beta_l)$ acts as a persistent channel for noise leakage. Through the identity mapping $x_{l+1} = x_l + \dots$, this leaked noise inevitably propagates and accumulates across the network depth, fundamentally compromising the discriminative boundary for generative artifacts. This analytic necessity motivates our transition from soft gradient suppression to explicit, hard layer pruning.

\subsection{Equivalence of Ridge Regression to Implicit Whitening and Activation Steering}
\label{app:ridge_equivalence}
In our binary calibration setting with mean-centered labels ($y \in \mathbb{R}^N$), the Ridge regression solver $(X_{\mathrm{global},s}^T X_{\mathrm{global},s} + \lambda I)^{-1} X_{\mathrm{global},s}^T y$ inherently unifies two critical mathematical operations:

\textbf{Implicit Whitening via Eigendecomposition.} The term $X_{\mathrm{global},s}^T X_{\mathrm{global},s}$ constitutes the empirical covariance matrix, which is real and symmetric. Applying the inverse of this matrix (or the regularized version $(X^T X + \lambda I)^{-1}$) is mathematically equivalent to performing an eigendecomposition $\Sigma = V \Lambda V^T$. Multiplying by the regularized precision matrix implicitly projects the features onto the orthogonal bases $V^T$ to strictly decorrelate them, and subsequently scales each dimension by the inverse of the shifted eigenvalues $(\Lambda + \lambda I)^{-1}$. Here, $\lambda$ prevents overzealous whitening from violently amplifying background noise.

\textbf{Activation Steering via Regularized Difference.} After the label mean-centering, the term $X_{\mathrm{global},s}^T y$ becomes strictly proportional to the difference of the class means $(\mu_1 - \mu_0)$. Consequently, the entire Ridge formulation mathematically simplifies to deriving a precision-scaled steering vector $w_s \propto \Sigma_s^{-1}(\mu_1 - \mu_0)$. Furthermore, injecting the Tikhonov regularization term ($\lambda I$) guarantees that the covariance matrix remains strictly invertible even in high-dimensional scenarios, enabling the stable extraction of the structural forgery residual.

\subsection{Semantic Soft-Routing Formulation}
\label{app:soft_routing}

Semantic Soft-Routing constructs virtual paired data $(x_{\mathrm{real}}^*, x_{\mathrm{fake}})$ from unpaired calibration samples, enabling the Ridge solver to isolate the forgery-specific residual. Let $X_r \in \mathbb{R}^{n_r \times d}$ and $X_f \in \mathbb{R}^{n_f \times d}$ denote the compressed real and fake feature manifolds, respectively.

\textbf{KNN-Based Soft Real Construction.} For each fake sample $x_f^{(i)}$, we retrieve its $K=5$ nearest real neighbors via cosine similarity in the $\ell_2$-normalized routing manifold $R$:
\begin{equation}
    \mathrm{sim}_{ij} = R_f^{(i)} \cdot R_r^{(j)}, \quad
    \mathcal{N}_K(i) = \mathrm{topK}(\mathrm{sim}_{i,:}, K{=}5)
\end{equation}
The soft real counterpart is constructed via temperature-scaled softmax blending with $\tau = 40.0$:
\begin{equation}
    w_{ij} = \frac{\exp(\mathrm{sim}_{ij} \cdot \tau)}{\sum_{j' \in \mathcal{N}_K(i)} \exp(\mathrm{sim}_{ij'} \cdot \tau)}, \quad
    x_{\mathrm{real}}^{*(i)} = \sum_{j \in \mathcal{N}_K(i)} w_{ij} \cdot x_r^{(j)}
\end{equation}

\textbf{Quality Gating and Quadratic Weighting.} To prevent low-quality pairings from contaminating the Ridge solution, we apply a quality filter with threshold $\delta = 0.70$:
\begin{equation}
    \mathcal{Q} = \{i \mid \max_j \mathrm{sim}_{ij} > \delta\}
\end{equation}
Filtered pairs receive a quadratic emphasis weight to prioritize high-confidence pairings:
\begin{equation}
    w_{\mathrm{pair}}^{(i)} = (\max_j \mathrm{sim}_{ij})^2, \quad \forall i \in \mathcal{Q}
\end{equation}
If fewer than 5 pairs pass the filter, we fall back to uniform weighting $w_{\mathrm{pair}}^{(i)} = 0.5$ to maintain numerical stability.

\textbf{Physical Meaning.} By synthesizing a virtual ``soft real'' sample as a convex combination of the most similar real neighbors for each fake sample, Semantic Soft-Routing converts the unpaired calibration set into approximately paired data $(x_{\mathrm{real}}^*, x_{\mathrm{fake}})$. This enables the Ridge classifier to compute a steering vector $w_s \propto \Sigma_s^{-1}(x_{\mathrm{real}}^* - x_{\mathrm{fake}})$ that isolates the \textit{forgery-specific residual} while canceling shared content and style variations. The quadratic weighting $\propto \mathrm{sim}^2$ ensures confident pairings dominate the solution, while a low-weight global mean stream provides a stable baseline. Without this routing, the linear solver would be forced to discriminate based on global distribution differences rather than the precise artifact residual.

\newpage
\section{Per-generator Balanced AP Comparison}
\label{app:ap_summary}

\begin{table}[h]
\centering
\caption{\textbf{Per-generator Balanced AP (\%) comparison on OpenFake.} All methods are trained or calibrated on FF++ (C23) and tested zero-shot on OpenFake. Best results in \textbf{bold}.}
\label{tab:openfake_per_gen_ap}
\resizebox{\textwidth}{!}{
\begin{tabular}{l|c|cccc|ccc|ccccc|c}
\toprule
\multicolumn{1}{c|}{\multirow{2}{*}{\textbf{Method}}} & \multirow{2}{*}{\makecell{\textbf{Trainable}\\ \textbf{Params}}} & \multicolumn{4}{c|}{\textbf{Stable Diffusion}} & \multicolumn{3}{c|}{\textbf{Proprietary}} & \multicolumn{5}{c|}{\textbf{Flow \& Emerging}} & \multirow{2}{*}{\textbf{Avg}} \\
\cmidrule(lr){3-6} \cmidrule(lr){7-9} \cmidrule(lr){10-14}
 & & \textbf{Sd-1.5} & \textbf{Sd-2.1} & \textbf{Sd-3.5} & \textbf{Sdxl} & \textbf{Mj-v6} & \textbf{Mj-v7} & \textbf{Dall-e 3} & \textbf{Flux.1} & \textbf{Flux-real} & \textbf{Ideogram} & \textbf{Recraft} & \textbf{Chroma} & \\
\midrule
Xception \cite{rossler2019faceforensics++} & 83M & 50.4 & 55.1 & 57.6 & 52.8 & 49.5 & 62.1 & 52.5 & 61.2 & 56.9 & 52.4 & 53.9 & 62.5 & 55.6 \\
F3Net \cite{qian2020thinking} & 22M & 39.7 & 49.2 & 55.4 & 49.0 & 44.0 & 55.2 & 44.9 & 56.4 & 54.9 & 48.3 & 53.6 & 55.0 & 50.5 \\
CORE \cite{ni2022core} & 22M & 37.7 & 45.7 & 52.1 & 50.2 & 42.2 & 57.2 & 44.8 & 59.3 & 58.0 & 50.7 & 48.8 & 53.3 & 50.0 \\
UCF \cite{yan2023ucf} & 47M & 43.1 & 48.0 & 54.8 & 52.5 & 44.4 & 48.5 & 42.4 & 58.1 & 56.5 & 47.8 & 51.2 & 55.0 & 50.2 \\
ProDet$^\dagger$ \cite{cheng2024can} & 96M & 49.1 & 52.8 & 52.6 & 57.1 & 52.9 & 65.6 & 51.1 & 60.0 & 57.8 & 51.4 & 61.7 & 57.7 & 55.8 \\
Effort$^\dagger$ \cite{yan2025orthogonal} & 0.19M & 69.8 & 72.2 & 51.8 & 61.4 & 49.5 & 54.6 & 44.4 & 46.0 & 47.1 & 50.2 & 55.0 & 51.7 & 54.5 \\
ForAda$^\dagger$ \cite{cui2025forensics} & 5.7M & 72.4 & 77.6 & 51.8 & 63.0 & 43.0 & 49.4 & 42.6 & 46.5 & 47.2 & 49.4 & 47.6 & 57.7 & 54.0 \\
GenD$^\dagger$ \cite{yermakov2026deepfake} & 0.03M & 76.4 & 80.7 & 56.1 & 68.0 & 49.2 & 58.5 & 61.1 & 46.9 & 47.3 & 54.8 & 50.9 & 57.6 & 59.0 \\
\midrule
\textbf{Lightning (Ours)} & \textbf{0M} & \textbf{78.5} & \textbf{90.8} & \textbf{82.7} & \textbf{88.0} & \textbf{63.6} & \textbf{85.7} & \textbf{67.4} & \textbf{85.0} & \textbf{81.4} & \textbf{77.8} & \textbf{69.7} & \textbf{88.2} & \textbf{79.9} \\
\bottomrule
\end{tabular}
}
\end{table}

\newpage
\counterwithin{table}{section}
\section{Full Per-Generator Results on OpenFake}
\label{app:full_results}

Table~\ref{tab:app_openfake_auc} and Table~\ref{tab:app_openfake_auc_2} report the complete per-generator AUC results across all 34 generators in OpenFake for all evaluated methods, while Table~\ref{tab:app_openfake_ap} and Table~\ref{tab:app_openfake_ap_2} report the corresponding Balanced AP results.

\begin{table}[h!]
\centering
\caption{\textbf{Complete per-generator AUC (\%) comparison on OpenFake (all 34 generators, part 1).}}
\label{tab:app_openfake_auc}
\resizebox{\textwidth}{!}{
\begin{tabular}{l|cccccccc}
\toprule
\textbf{Generator} & \textbf{Xception} & \textbf{F3Net} & \textbf{CORE} & \textbf{UCF} & \textbf{ProDet} & \textbf{Effort} & \textbf{GenD} & \textbf{Lightning} \\
\midrule
sd-2.1 & 0.590 & 0.492 & 0.450 & 0.499 & 0.534 & 0.739 & 0.834 & \textbf{0.905} \\
chroma & 0.637 & 0.585 & 0.548 & 0.597 & 0.585 & 0.534 & 0.598 & \textbf{0.887} \\
sdxl & 0.515 & 0.473 & 0.512 & 0.523 & 0.566 & 0.635 & 0.711 & \textbf{0.886} \\
frames-23-1-25 & 0.565 & 0.575 & 0.506 & 0.600 & 0.633 & 0.635 & 0.629 & \textbf{0.883} \\
ideogram-2.0 & 0.549 & 0.436 & 0.463 & 0.479 & 0.563 & 0.628 & 0.616 & \textbf{0.881} \\
flux.1-schnell & 0.598 & 0.566 & 0.564 & 0.599 & 0.606 & 0.515 & 0.556 & \textbf{0.869} \\
midjourney-7 & 0.598 & 0.555 & 0.577 & 0.477 & 0.660 & 0.587 & 0.628 & \textbf{0.864} \\
flux.1-dev & 0.563 & 0.585 & 0.620 & 0.609 & 0.608 & 0.440 & 0.474 & \textbf{0.855} \\
halfmoon-4-4-25 & 0.630 & 0.631 & 0.609 & 0.647 & 0.668 & 0.603 & 0.649 & \textbf{0.848} \\
sd-3.5 & 0.635 & 0.588 & 0.523 & 0.595 & 0.538 & 0.521 & 0.591 & \textbf{0.844} \\
flux-realism & 0.529 & 0.567 & 0.615 & 0.581 & 0.584 & 0.468 & 0.472 & \textbf{0.830} \\
ideogram-3.0 & 0.535 & 0.474 & 0.526 & 0.475 & 0.521 & 0.510 & 0.567 & \textbf{0.816} \\
hidream-i1-full & 0.661 & 0.591 & 0.551 & 0.627 & 0.584 & 0.461 & 0.542 & \textbf{0.814} \\
sdxl-touchofrealism & 0.470 & 0.513 & 0.540 & 0.490 & 0.555 & 0.562 & 0.665 & \textbf{0.812} \\
sd-1.5 & 0.562 & 0.348 & 0.312 & 0.423 & 0.543 & 0.718 & 0.798 & \textbf{0.800} \\
sd-1.5-dreamshaper & 0.339 & 0.325 & 0.426 & 0.268 & 0.411 & 0.661 & 0.729 & \textbf{0.787} \\
sdxl-realvis-v5 & 0.476 & 0.428 & 0.401 & 0.392 & 0.400 & 0.513 & 0.639 & \textbf{0.784} \\
\bottomrule
\end{tabular}
}
\end{table}

\begin{table}[h!]
\centering
\caption{\textbf{Complete per-generator AUC (\%) comparison on OpenFake (all 34 generators, part 2).}}
\label{tab:app_openfake_auc_2}
\resizebox{\textwidth}{!}{
\begin{tabular}{l|cccccccc}
\toprule
\textbf{Generator} & \textbf{Xception} & \textbf{F3Net} & \textbf{CORE} & \textbf{UCF} & \textbf{ProDet} & \textbf{Effort} & \textbf{GenD} & \textbf{Lightning} \\
\midrule
lumina-17-2-25 & 0.767 & 0.735 & 0.683 & 0.690 & 0.678 & 0.615 & 0.633 & \textbf{0.776} \\
sd-1.5-epicdream & 0.455 & 0.493 & 0.483 & 0.461 & 0.493 & 0.554 & 0.647 & \textbf{0.773} \\
mystic & 0.539 & 0.562 & 0.621 & 0.580 & 0.534 & 0.574 & 0.536 & \textbf{0.773} \\
recraft-v3 & 0.572 & 0.575 & 0.517 & 0.524 & 0.625 & 0.580 & 0.543 & \textbf{0.748} \\
flux-amateursnapshotphotos & 0.575 & 0.618 & 0.610 & 0.627 & 0.585 & 0.518 & 0.453 & \textbf{0.741} \\
aurora-20-1-25 & 0.538 & 0.590 & 0.524 & 0.496 & 0.709 & 0.638 & 0.626 & \textbf{0.727} \\
dalle-3 & 0.564 & 0.463 & 0.455 & 0.419 & 0.462 & 0.432 & 0.664 & \textbf{0.706} \\
sdxl-juggernaut & 0.582 & 0.537 & 0.492 & 0.572 & 0.461 & 0.502 & 0.568 & \textbf{0.705} \\
imagen-4.0 & 0.647 & 0.549 & 0.509 & 0.567 & 0.523 & 0.494 & 0.550 & \textbf{0.694} \\
midjourney-6 & 0.536 & 0.446 & 0.405 & 0.455 & 0.539 & 0.518 & 0.531 & \textbf{0.693} \\
gpt-image-1 & 0.534 & 0.496 & 0.514 & 0.473 & 0.414 & 0.519 & 0.549 & \textbf{0.666} \\
recraft-v2 & 0.604 & 0.588 & 0.479 & 0.559 & 0.559 & 0.619 & 0.530 & \textbf{0.657} \\
sdxl-epic-realism & 0.534 & 0.534 & 0.481 & 0.477 & 0.466 & 0.514 & 0.605 & \textbf{0.643} \\
imagen-3.0-002 & 0.608 & 0.527 & 0.468 & 0.538 & 0.468 & 0.524 & 0.503 & \textbf{0.633} \\
flux-1.1-pro & 0.564 & 0.533 & 0.495 & 0.563 & 0.584 & 0.515 & 0.468 & \textbf{0.618} \\
grok-2-image-1212 & 0.452 & 0.505 & 0.527 & 0.430 & \textbf{0.633} & 0.479 & 0.478 & 0.576 \\
flux-mvc5000 & 0.515 & 0.534 & 0.504 & 0.551 & \textbf{0.559} & 0.432 & 0.372 & 0.446 \\
\bottomrule
\end{tabular}
}
\end{table}

\begin{table}[h!]
\centering
\caption{\textbf{Complete per-generator Balanced AP (\%) comparison on OpenFake (all 34 generators, part 1).}}
\label{tab:app_openfake_ap}
\resizebox{\textwidth}{!}{
\begin{tabular}{l|cccccccc}
\toprule
\textbf{Generator} & \textbf{Xception} & \textbf{F3Net} & \textbf{CORE} & \textbf{UCF} & \textbf{ProDet} & \textbf{Effort} & \textbf{GenD} & \textbf{Lightning} \\
\midrule
sd-2.1 & 55.1 & 49.2 & 45.7 & 48.1 & 52.8 & 72.2 & 80.6 & \textbf{90.8} \\
chroma & 62.4 & 55.0 & 53.3 & 55.0 & 57.7 & 51.7 & 57.6 & \textbf{88.1} \\
sdxl & 52.8 & 49.0 & 50.1 & 52.6 & 57.1 & 61.4 & 68.0 & \textbf{88.0} \\
frames-23-1-25 & 53.8 & 54.4 & 50.0 & 58.5 & 59.0 & 58.4 & 57.3 & \textbf{87.1} \\
ideogram-2.0 & 51.6 & 43.9 & 46.1 & 47.8 & 55.4 & 58.5 & 56.6 & \textbf{86.4} \\
flux.1-schnell & 61.1 & 54.4 & 55.7 & 55.4 & 58.5 & 50.6 & 55.0 & \textbf{86.8} \\
midjourney-7 & 62.1 & 55.2 & 57.2 & 48.6 & 65.6 & 54.6 & 58.5 & \textbf{85.6} \\
flux.1-dev & 61.3 & 56.4 & 59.2 & 58.2 & 60.0 & 46.0 & 46.9 & \textbf{85.0} \\
halfmoon-4-4-25 & 64.1 & 62.1 & 60.7 & 63.9 & 67.1 & 55.7 & 58.6 & \textbf{83.1} \\
sd-3.5 & 57.6 & 55.4 & 52.0 & 54.9 & 52.6 & 51.8 & 56.1 & \textbf{82.7} \\
flux-realism & 56.9 & 54.9 & 58.0 & 56.6 & 57.8 & 47.1 & 47.3 & \textbf{81.4} \\
ideogram-3.0 & 52.4 & 48.3 & 50.6 & 47.9 & 51.4 & 50.2 & 54.8 & \textbf{77.8} \\
hidream-i1-full & 64.2 & 55.0 & 54.3 & 57.7 & 57.1 & 46.9 & 51.5 & \textbf{78.9} \\
sdxl-touchofrealism & 51.0 & 52.6 & 52.1 & 51.5 & 57.3 & 55.0 & 64.0 & \textbf{77.8} \\
sd-1.5 & 50.4 & 39.7 & 37.7 & 43.1 & 49.1 & 69.8 & 76.4 & \textbf{78.5} \\
sd-1.5-dreamshaper & 39.0 & 40.3 & 42.1 & 37.6 & 43.5 & 62.1 & 67.4 & \textbf{74.1} \\
sdxl-realvis-v5 & 45.6 & 43.5 & 41.4 & 42.7 & 43.2 & 51.0 & 60.7 & \textbf{75.5} \\
\bottomrule
\end{tabular}
}
\end{table}

\begin{table}[h!]
\centering
\caption{\textbf{Complete per-generator Balanced AP (\%) comparison on OpenFake (all 34 generators, part 2).}}
\label{tab:app_openfake_ap_2}
\resizebox{\textwidth}{!}{
\begin{tabular}{l|cccccccc}
\toprule
\textbf{Generator} & \textbf{Xception} & \textbf{F3Net} & \textbf{CORE} & \textbf{UCF} & \textbf{ProDet} & \textbf{Effort} & \textbf{GenD} & \textbf{Lightning} \\
\midrule
lumina-17-2-25 & 75.5 & 71.1 & 70.8 & 59.3 & 63.6 & 56.6 & 56.3 & \textbf{78.8} \\
sd-1.5-epicdream & 47.6 & 48.7 & 46.0 & 50.2 & 53.6 & 53.3 & 61.8 & \textbf{74.8} \\
mystic & 58.0 & 57.4 & 61.9 & 55.9 & 54.6 & 55.4 & 52.5 & \textbf{73.3} \\
recraft-v3 & 53.8 & 53.6 & 48.9 & 51.3 & 61.7 & 55.0 & 50.9 & \textbf{69.7} \\
flux-amateursnapshotphotos & 59.4 & 59.9 & 59.7 & 58.2 & 55.9 & 51.6 & 45.9 & \textbf{71.8} \\
aurora-20-1-25 & 53.5 & 54.4 & 50.1 & 50.2 & \textbf{68.1} & 60.8 & 56.9 & 67.2 \\
dalle-3 & 52.5 & 44.9 & 44.8 & 42.5 & 51.1 & 44.5 & 61.1 & \textbf{67.4} \\
sdxl-juggernaut & 56.0 & 53.3 & 49.8 & 52.1 & 48.3 & 49.8 & 55.2 & \textbf{65.2} \\
imagen-4.0 & 59.9 & 52.3 & 51.2 & 50.7 & 50.2 & 48.4 & 51.0 & \textbf{62.8} \\
midjourney-6 & 49.5 & 44.0 & 42.2 & 44.5 & 52.9 & 49.5 & 49.2 & \textbf{63.6} \\
gpt-image-1 & 53.7 & 48.9 & 50.0 & 47.8 & 46.7 & 50.8 & 52.3 & \textbf{68.5} \\
recraft-v2 & 54.7 & 53.3 & 48.8 & 51.7 & 55.0 & 58.6 & 48.9 & \textbf{59.2} \\
sdxl-epic-realism & 52.3 & 52.8 & 46.9 & 49.4 & 49.4 & 50.7 & 58.3 & \textbf{59.6} \\
imagen-3.0-002 & 55.9 & 50.6 & 47.5 & 49.6 & 47.0 & 51.1 & 47.7 & \textbf{57.4} \\
flux-1.1-pro & 57.2 & 52.4 & 49.7 & 55.3 & 56.7 & 51.0 & 46.6 & \textbf{58.2} \\
grok-2-image-1212 & 48.1 & 48.6 & 49.8 & 45.4 & \textbf{64.2} & 47.2 & 48.2 & 52.5 \\
flux-mvc5000 & 52.2 & 52.1 & 49.6 & 54.2 & 54.6 & 45.5 & 41.0 & \textbf{45.4} \\
\bottomrule
\end{tabular}
}
\end{table}

\newpage
\section{Decoupled Augmentation for Anomaly Manifold Calibration}
\label{app:decoupled_aug}
Unlike over-parameterized deep neural networks that benefit uniformly from data augmentation, rigid statistical solvers (e.g., Ridge Regression) are highly susceptible to covariance pollution. As established in robust statistics \citep{huber2009robust, ledoit2004well}, injecting contaminated or noisy samples directly distorts the empirical covariance matrix and fundamentally degrades its condition number. For high-dimensional estimators, this intrinsic ill-conditioning makes the analytic solution highly unreliable, effectively drowning out subtle, discriminative artifact signals and leading to catastrophic underfitting. However, completely abandoning augmentation sacrifices potential robustness against unseen perturbations.

To circumvent this dilemma, we propose a decoupled paradigm. Rather than polluting the analytic hyperplane, we isolate the augmented samples exclusively to construct an independent ``Augmented Anchor''. By calculating the precision matrix of this anchor, we measure the Mahalanobis distance $d$ of any query sample to the natural manifold. We then introduce an exponential decay penalty:
\begin{equation}
    \sigma = 1 + \alpha \exp(-\gamma \cdot d^2)
\end{equation}
During inference, the raw linear prediction is modulated by dividing it by $\sigma$. This elegantly transforms data augmentation from a risky generative regularizer into a safe, strict anomaly penalty against OOD attacks, successfully bridging the gap between statistical solvers and data augmentation without compromising the analytic boundary.

\section{Federated Covariance Aggregation Details}
\label{app:federated}
In a decentralized network across $M$ edge clients, the global covariance elegantly decomposes via strict mathematical addition: $(X^T X)_{\mathrm{global}} = \sum_{i=1}^M X_i^T X_i$. Rather than transmitting massive gradient updates, clients locally extract and transmit merely a negligible $K \times K$ covariance matrix (e.g., a mere $\sim$8 KB payload in FP16 for $K=64$). This strict decomposition provides a strong privacy safeguard, as raw image manifolds cannot be trivially reconstructed from aggregated statistical moments, while operating at near-zero network I/O, uniquely enabling secure, instantaneous federated collaboration against emerging threats.

\end{document}